\documentclass{bmvc2k}
\usepackage{gensymb}
\usepackage{mathtools}
\usepackage{amsmath}
\usepackage{graphicx}
\usepackage{float}
\usepackage{subfigure}

\DeclarePairedDelimiter{\norm}{\lVert}{\rVert}

\title{RankPose: Learning Generalised Feature\\ with Rank Supervision for Head Pose Estimation}

\addauthor{Donggen Dai}{daidonggen@gmail.com}{1}
\addauthor{Wangkit Wong}{wwongaa@connect.ust.hk}{1}
\addauthor{Zhuojun Chen}{chenzhuojun373@pingan.com.cn}{1}

\addinstitution{
 Ping An Technology\\
 Shenzhen, China
}

\runninghead{~}{~}

\begin{document}

\maketitle

\begin{abstract}
We address the challenging problem of RGB image-based head pose estimation. We first reformulate head pose representation learning to constrain it to a bounded space. Head pose represented as vector projection or vector angles shows helpful to improving performance. Further, a ranking loss combined with MSE regression loss is proposed.
The ranking loss supervises a neural network with paired samples of the same person and penalises incorrect ordering of pose prediction. Analysis on this new loss function suggests it contributes to a better local feature extractor, where features are generalised to Abstract Landmarks which are pose-related features instead of pose-irrelevant information such as identity, age, and lighting.
Extensive experiments show that our method significantly outperforms the current state-of-the-art schemes on public datasets: AFLW2000 and BIWI.
Our model achieves significant improvements over previous SOTA MAE on AFLW2000 and BIWI from 4.50~\cite{hsu2018quatnet} to 3.66 and from 4.0~\cite{yang2019fsa} to 3.71 respectively.
Source code will be made available at: 

https://github.com/seathiefwang/RankHeadPose.

\end{abstract}

\section{Introduction}
\label{sec:intro}
Head pose estimation has aroused much interest recently as it serves as auxiliary information improving the accuracy of many computer vision tasks such as face recognition, expression recognition and attention analysis.
It aims at predicting mathematical representations for human head orientation (e.g. Euler angles, quaternion~\cite{hsu2018quatnet}). There exist accurate solutions~\cite{meyer2015robust, ding2015robust} that estimate head pose from depth images which are, however, not accessible in many real-world applications.
Hence, building a model based on 2D images has emerged as a good trade-off between complexity and accuracy. There are two-stage methods~\cite{hartley2003multiple} that predict facial landmarks first to approach head pose estimation. The accuracy of the second stage, i.e. head pose estimation, is heavily constrained by the first stage landmark prediction which compromises under conditions of poor lighting and large head pose. Considering that, DCNN methods
based on dense face alignment and multi-task learning for estimating both facial landmarks and head pose are recently proposed~\cite{zhu2017face, feng2018joint, ranjan2017hyperface}.
But, such multi-objective design has not focused on head pose, resulting in suboptimal pose estimation~\cite{ruiz2018fine}. Instead, dedicated head pose estimation models combining L2 regression loss and specifically designed losses achieve superior performance over landmark-based schemes ~\cite{ruiz2018fine, yang2019fsa}.

In this paper, we conduct an in-depth study on 2D image head pose estimation. We design a novel deep learning approach to seek to improve current methods.
Firstly, we reformulate the output function of a regression neural network in a normalisation manner to achieve a model closer to the head pose representation. In light of that Euler angles are bounded (within $[-\pi/2, \pi/2]$), we as well bound the model output space by L2 normalisation of deep head pose features $F$ and the weight vectors $W$ of the output layer. This results in a cosine transformation with which head pose is predicted as the projection of $F$ on $W$ (also known as cosine similarity between $F$ and $W$). Further, we apply an inverse cosine function (Arccos) to arrive at an angle model specialised for Euler angle representation. By so doing, head pose is estimated as the angle between $F$ and $W$ . Our experimental results show this approach has substantial positive effect on head pose prediction accuracy.

More importantly, we introduce a new loss function for head pose estimation.
We address the fact that a face image consists of two major components: global and local features. Pose-related features which a good model is expected to extract should be invariant with most of the global features such as lighting condition, biological appearance, facial expression, etc.
Motivated by this, we employ a Siamese architecture and a novel loss function combining MSE regression loss and a pairwise ranking loss.
The pairwise ranking loss takes two images of the same person as input and requires the network to rank angles (yaw, pitch, and roll) of them so as to predict larger values for the image with larger ground-truth label.
Since for most training data, images of the same person are obtained under similar conditions, the neural network learns from pairs of inputs saturated with global similarities (pose-irrelevant futures such as identity and lighting) yet only differentiable by local features (e.g. nose shape, pupillary distance). In such cases, ordering of head pose angles highlights pose-related features.
The ranking loss thus penalises incorrect ordering of head pose predictions of similar input images and the learnt model can be more generalisable to condition changes. We analytically show that local features which we consider as abstract landmarks are crucial to accurate prediction. We name our method RankPose.

The main contributions of this paper are summarised as follows.
\begin{itemize}
  \item We propose a reformulation of head pose representation learning by adopting a transformation scheme. We show that transforming model output to angle space helps improve head pose prediction accuracy.
    \item We propose a novel CNN model exploring Siamese architecture and ranking loss to distinguish pose-related features from a mixture of pose-related and -irrelevant features.
  \item We improve the current state-of-the-art results on public benchmarks: AFLW2000 and BIWI significantly.
        Our model improves the MAE on AFLW2000 and BIWI from 4.50~\cite{hsu2018quatnet} to 3.66 and from 4.0~\cite{yang2019fsa} to 3.71 respectively.
\end{itemize}

\section{Related Work}
\label{sec:related}
In this section, we briefly revisit previous works of head pose estimation. Previous methods used as comparison schemes in the Experimental section~\ref{results} are emphasised with bold text.

\emph{Methods using depth information}:
Martin et. al.~\cite{martin2014real} propose an iterative closest point algorithm to register depth image with a 3D head model.
Initial pose is estimated using features of the head, and then the algorithm improves the current estimated pose over time.
A similar method is proposed in~\cite{meyer2015robust}, which combines particle swarm optimisation and the iterative closest point algorithm to eliminate the need for explicit initialisation.
Random regression forests trained on synthetic data for head pose estimation are presented in~\cite{fanelli2011real}.
Mukherjee et. al.~\cite{mukherjee2015deep} present a CNN-based model for head pose estimation running on RGB-D data, which is first trained to predict coarse gazing direction classes and then fine-tuned to regress head poses.  Though these methods achieve accurate estimation, depths data are not accessible in many real-world applications.


\emph{Landmark-based methods}:
Early landmark-based works use the Perspective-n-Point (PnP) algorithm~\cite{hartley2003multiple} to estimate pose from a single RGB image.
The PnP algorithm is 2-step where it first obtains 2D facial landmarks, and then establishes correspondence between the 2D landmarks and a generic 3D head model.
Subsequently, face alignment is performed on the correspondence between the 2D landmarks and the 3D head pose is obtained from the rotation matrix.
\textbf{Dlib} presented in~\cite{kazemi2014one} uses an ensemble of regression
trees to localise facial landmarks with realtime prediction speed.
Recently, methods that estimate 2D facial landmarks using modern deep learning networks~\cite{bulat2017far, gupta2019nose} have become dominant.
\textbf{KEPLER}~\cite{kumar2017kepler} predicts landmarks and pose simultaneously using an improved GoogLeNet architecture.
Landmark localisation is improved by coarse pose supervision.
\textbf{Two-Stage}~\cite{lv2017deep} uses a deep regression network with two-stage re-initialisation for initial landmark localisation.
However, the performance of sparse facial landmark localisation for face alignment often drops under large head pose and heavy occlusion.
To address these issues, dense face alignment methods~\cite{feng2018joint, zhu2016face} based on deep CNNs have been proposed.
\textbf{3DDFA}~\cite{zhu2016face} uses a cascaded CNN to fit a dense 3D face model to face images for face alignment.
Head pose is obtained as a by-product of 3D face alignment.
Posing facial feature analysis as a multi-task problem, Hyperface~\cite{ranjan2017hyperface}
learns a shared CNN to perform face detection, facial landmark localisation, head pose estimation, and gender recognition simultaneously.
To improve landmark localisation accuracy for low-resolution images, \textbf{FAN}~\cite{bulat2018super} simultaneously improves face resolution and detects the facial landmarks.

\emph{Landmark-free methods}:
Models designed for landmark localisation or dense face alignment may not have focused on head pose accuracy.
Instead, landmark-free approaches have emerged as more promising schemes.
Ruiz et. al.~\cite{ruiz2018fine} propose \textbf{HopeNet}, a multi-loss CNN trained on synthetic data to predict intrinsic Euler angles from image intensities.
The classification loss guides the networks to learn the ordering of neighbourhood of poses in a robust fashion.
It leverages feature aggregation across different semantic levels using a fine-grained spatial structure and soft stagewise regression.
\textbf{FSA-Net}~\cite{yang2019fsa} achieves state-of-the-art accuracy by adopting an attention mechanism.
To avoid the ambiguity problem in the commonly used Euler angle representation,
\textbf{QuatNet}~\cite{hsu2018quatnet} trains ResNet50 to predict the quaternion of head pose instead of Euler angles.
QuatNet adopts a multi-regression loss function which addresses the non-stationary property in head pose estimation by ranking different intervals of the angles with a classification loss.

It is worth noting that our ranking supervision scheme is fundamentally different from QuatNet's in two aspects: 1) Our model utilises Siamese architecture to measure ranking errors between paired samples to focus on pose-related features while QuatNet puts ranking as more of a classification problem; 2) QuatNet aims at tackling the non-stationary property, whereas our RankPose makes effort to differentiate local changes.

\section{Method}
\label{sec:method}
In this section, we introduce two key ingredients of the proposed RankPose in detail.
The output space transformation is described in Section~\ref{arccosine}.
Our loss function is described in Section~\ref{siamese}. In Section~\ref{visual}, we provide an abstract landmark view to interpret our RankPose method.

\subsection{Output Space Transformation}
\label{arccosine}
Most commonly, a CNN-based head pose regression model takes input image $X$ and predicts head pose representation $Y$. Dissecting the network to a DCNN backbone ($D$) and the final output layer (FC), $Y$ is predicted as dot product of a feature vector $F=D(X)$ produced by the backbone and the weight matrix $W$ in FC:
\begin{align}
Y_{dot} = W^{T}F. \label{dot}
\end{align}

In the above equation, $Y_{dot}$ is unbounded ($Y_{dot}^{d} \in R$) but head pose representation should not. Euler angles, for instance, are bounded within $[-\pi/2, \pi/2]$ in this case. To rectify, we adopt a transformation scheme.

We first apply L2 normalisation to $F$ and $W$, s.t. $\norm{F}=1$ and $\norm{W_{j}}=1$.
Vector $F$ and each column $j$ of $W$ are normalised to have unit length.
The L2 normalised version of $F$ and $W$ are denoted as $\hat{F}$ and $\hat{W}$ respectively.
In this way, $Y$ is obtained as projections of $\hat{F}$ on $\hat{W_{j}}$ which is also the cosine similarity between $\hat{F}$ and $\hat{W_{j}}$ since $Y_{j}=\alpha(\norm{F}\norm{W_{j}}cos\theta_{j})=\alpha cos\theta_{j}$, so it becomes:
\begin{align}
Y_{cosine} =\alpha \hat{W}^{T}\hat{F}, \label{cosine}
\end{align}
where $\alpha$ is a scalar that scales the output range from $[-1, 1]$ to the head pose range. Here, $\alpha=\pi/2$ for Euler angles.

The above is the Cosine transformation which transforms the model output to
the range of cosine function. Moreover, for Euler angle representation, substitution of $\alpha$ with the inverse cosine function ($Arccos$) changes the measurement into the geometric angle $\theta_{j}$ between $\hat{F}$ and
$\hat{W_{j}}$ since $Y_{j}=Arccos(cos\theta_{j})=|\theta_{j}|$ where, for $\theta_{j} \in [0,\pi]$, the Euler angles are linearly associated with the angles between $\hat{F}$ and
$\hat{W_{j}}$. To obtain Euler angles in range $[-\pi/2,\pi/2]$, a $\pi/2$ shift to the function is further performed, and $Y$ becomes:
\begin{align}
Y_{arccos} =Arccos(\hat{W}^{T}\hat{F})-\pi/2. \label{arccos}
\end{align}

\subsection{Pairwise Ranking Loss}
\label{siamese}

\begin{figure}
    \centering
    \includegraphics[width=0.6 \textwidth]{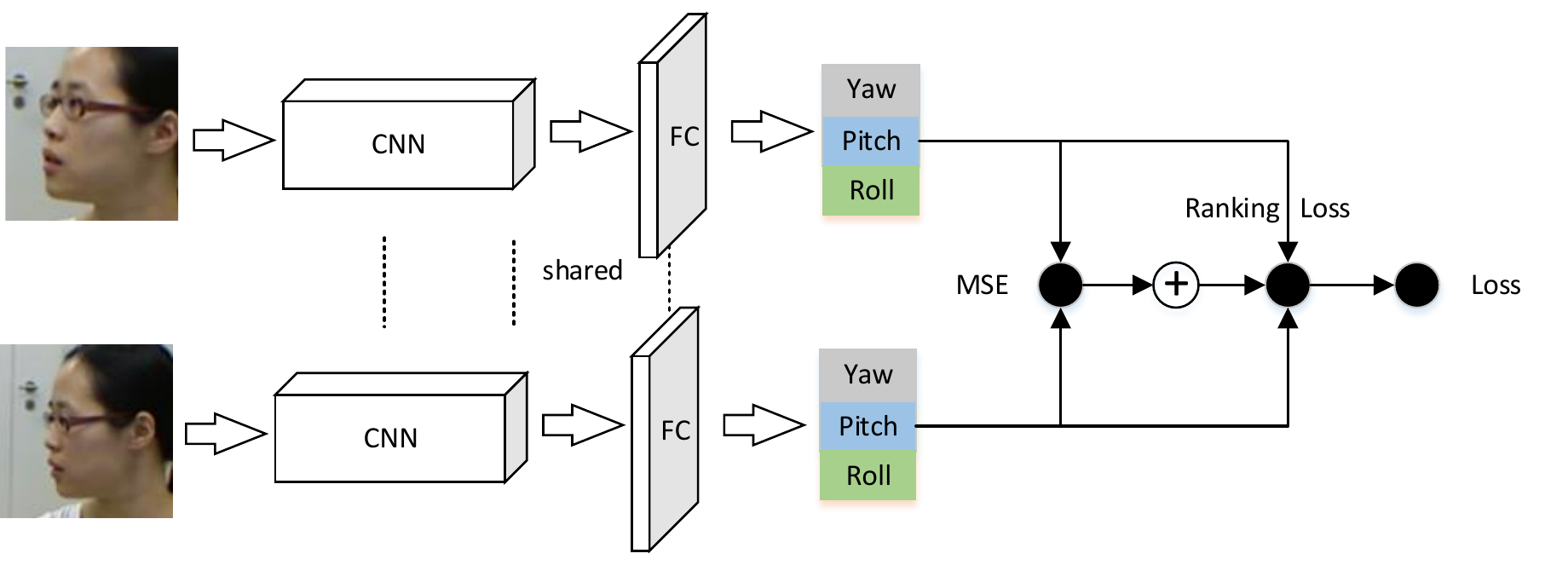}
    \caption{Illustration of our proposed model: Two images of the same person are passed through a CNN model, which predicts $3$
angle values representing pitch, yaw, and roll respectively. The prediction is supervised with a loss function combining MSE loss and ranking loss.}
    \label{fig:siamese}
\end{figure}

As shown in Figure~\ref{fig:siamese}, the model consists of a CNN backbone. The backbone can be one of
the modern deep convolutional neural networks such as VGG~\cite{simonyan2014very}, GoogleNet~\cite{szegedy2015going} or ResNet~\cite{he2016deep}. The network forwards two images and always computes loss function of both at the same time, and in the backpropagation, the two images condition each other. This is known as the Siamese architecture~\cite{bromley1994signature}.
In a training iteration, the network is trained with a mini-batch of $N$ training samples.
Each training sample $\{X_{i1}, X_{i2}\}$ is a pair of sampled training images of the same person, where $i$
is the sample index. Subsequently, $\{Y_{i1}, Y_{i2}\}$ is the head pose prediction and $\{\tilde{Y}_{i1}, \tilde{Y}_{i2}\}$ is the ground-truth head pose. For Euler angles in our case, $Y$ is a three dimensional vector representing yaw, pitch and roll respectively.

In most regression models, mean squared error (MSE) is a close and effective loss function for modelling minimisation problem of a particular dataset.
In our model, it is also necessary to make the predictions match the real head pose for any given image as much as possible.
Therefore, one key objective is the following:
\begin{align}
 L_{MSE} = \frac{1}{N} \sum_{i=1}^{N}  \left( \norm{\tilde{Y}_{i1}-Y_{i1} }^{2} + \norm{\tilde{Y}_{i2}-Y_{i2} }^{2}  \right) \label{mse_loss},
\end{align}
where $\norm{.}$ is the L2 norm. However, it is well-known that MSE is prone to overfit to outliers because it heavily weights them. As the model moves towards outliers, it learns pose-irrelevant features to memorise the dataset which makes it lack generalisation.
To learn generalised pose-related features which are invariant to biological appearance and lighting condition etc., we impose the following pairwise ranking loss on the model.
\begin{align}
 L_{ranking} =  \frac{1}{N} \sum_{i=1}^{N} \sum_{d} \left[ \max  \left\{ 0,   - sgn{(\tilde{Y}_{i2}-\tilde{Y}_{i1} )}  \cdot (Y_{i2}-Y_{i1}) \right\} \right]^{d}  \label{rank_loss},
\end{align}
where $sgn(z)$ is the sign function that returns $1$ if $z$ is greater than $0$, returns $-1$ if $z$ is less than $0$, and returns $0$ otherwise.
Operation $\{\cdot\}$ is the element-wise product between vectors.
$\left[v\right]^{d}$ is the $d$th element of vector $v$ and $\sum_{d}\left[v\right]^d$ sums all elements of vector $v$.
$\max  \left\{ 0,   - sgn{(\tilde{Y}_{i2}-\tilde{Y}_{i1} )}  \cdot (Y_{i2}-Y_{i1}) \right\}$ is a 3-dimensional vector, whose elements
represent the pairwise ranking loss of yaw, pitch, and roll respectively.
Its $d$th element is,
\begin{align}
\left[ \max  \left\{ 0,   - sgn{(\tilde{Y}_{i2}-\tilde{Y}_{i1} )}  \cdot (Y_{i2}-Y_{i1}) \right\} \right]^{d} = \begin{cases}
                         \max \left\{ 0,   Y_{i2}^{d}-Y_{i1}^{d}  \right\}   &  \mbox{if}~~ \tilde{Y}_{i2}^{d}\leq \tilde{Y}_{i1}^{d} \\
                         \max \left\{ 0,   Y_{i1}^{d}-Y_{i2}^{d}  \right\}   &  \mbox{if}~~ \tilde{Y}_{i1}^{d}\leq \tilde{Y}_{i2}^{d}
                       \end{cases},
\end{align}
which requires the prediction of yaw (or pitch or roll) angle, $Y_{i1}^{d}$ of image $X_{i1}$ to be larger than the prediction of
$Y_{i2}^{d}$, if the yaw (or pitch or roll) ground-truth label $\tilde{Y}_{i1}^{d}$ of image $X_{i1}$ is larger.
We supervise the network with a novel loss function combining MSE loss and pairwise ranking loss, given by:
\begin{align}
L = \beta L_{MSE} + (1-\beta) L_{ranking} \label{loss},
\end{align}
where $\beta$ trades off the importance between the MSE loss and the ranking loss.

\vspace{-0.5cm}
\subsection{Visualisation $\&$ Abstract Landmark Interpretation}
\label{visual}

\begin{figure}
\centering
\includegraphics[width=0.5\textwidth]{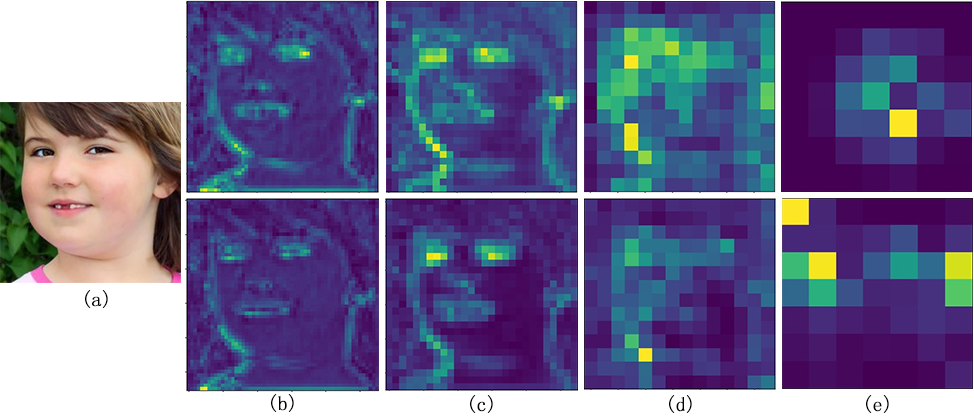}
\setlength{\belowcaptionskip}{-8pt}
\caption{An input image and its corresponding activation maps from the first, second and third residual blocks. The upper row is produced by RankPose, the lower being of a model learnt with MSE loss.}
\label{fig:act_map}
\end{figure}

Facial landmark-based methods rely purely on local conditioning features, i.e. the exact landmark locations. Their success suggests that local information is good representation for head pose. Efficient as they remain in practice, landmark-based methods appear to underwhelmingly perform when a pose regression CNN is a feasible option. Needless to say, CNN can learn strong abstract features, but we observe that an accurate model tends to be more confined to local features such as the face silhouette and edges of eyes and mouth. We consider those features as the abstract landmarks which can be much more informative than the explicit landmark coordinates.

Figure~\ref{fig:act_map} visualises the activation maps learnt by RankPose.
Activation maps from the first, second, and third residual blocks of ResNet50 are visualised in the top row of Fig.(b), Fig.(c), Fig.(d) and Fig.(e) respectively.
Correspondingly, the activation maps of a model trained with MSE loss without using transformation are visualised in the bottom row.
Activation values are normalised to $[0, 1]$. Brighter colours specify greater values.
It can be seen from the figure that RankPose produces sharper edges compared to the MSE model.
RankPose learns stronger activation near abstract facial landmarks.
We believe that facial landmarks are generalisable features for head pose estimation, which justifies that RankPose learns more generalised features.

\begin{figure}
\centering
\includegraphics[width=0.5 \textwidth]{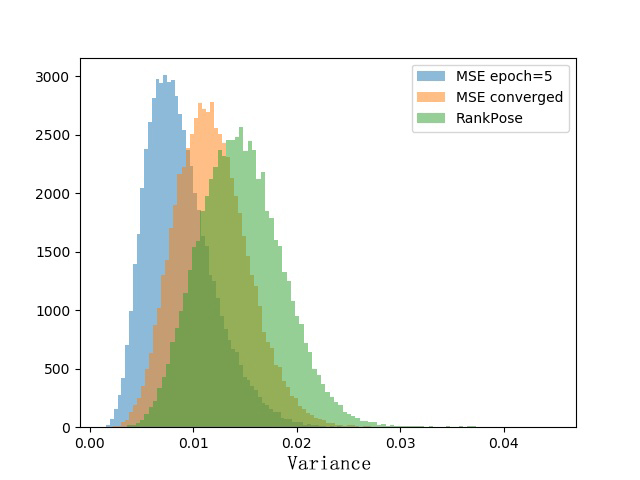}
\setlength{\belowcaptionskip}{-8pt}
\caption{Distribution of variance of the activation maps produced by the first residual block}
\label{fig:variance_dist}
\end{figure}

Figure~\ref{fig:variance_dist} shows the variance of the activation maps produced by the first residual block.
Variance is computed as $\sum_{i=1}^{n}\frac{1}{n}(pix_{i}-\overline{pix})^{2} $, where $pix_{i}$ is the value of the $i$th pixel of the activation map
and $\overline{pix}$ is the mean value.
The peak in blue colour is of an MSE model at an early training stage. Keeping training the model to convergence results in the peak in salmon colour. The peak in green where there is the highest variance distribution is of RankPose. Better models exhibit clearly different statistics over activation variances. Higher variance indicates a better local feature extractor.
It suggests RankPose learns more discriminative features.

\section{Experiments}
\label{sec:epx}
In this section, we present implementation, datasets, evaluation protocols, benchmark results and ablation studies.
\vspace{-0.5cm}
\subsection{Implementation}
To make fair comparison with previous works, we build our model on top of a ResNet50 backbone pre-trained on ImageNet.
The last FC layer is replaced by new output layer with weights of shape $[2048, 3]$.
Unless otherwise stated, as shown in Section~\ref{arccosine}, each column of the weight matrix is normalised to have unit length.
Random cropping, random scale, Gaussian noise, and Gaussian blur are adopted as standard augmentation pipeline.
The model is trained by Adam optimisation for $80$ epochs, with an initial learning rate of $0.001$ and $128$ batch size.
Cosine decay~\cite{loshchilov2016sgdr} is applied to the learning rate during training.
In all experiments, we sample randomly two images of the same identity to form a training sample $\{X_{i1}, X_{i2}\}$.
$\beta$ in the loss function (Equation~\ref{loss}) is set to 0.5.

\vspace{-0.5cm}
\subsection{Datasets \& Evaluation Protocols}
\emph{Training dataset}: RankPose is trained on 300W-LP~\cite{zhu2016face}. 300W-LP is synthesised from 300W dataset by a 3DMM model.
The synthesis is done by predicting the depth of face image.
Then, the profile views are generated by 3D rotation of the model.


\emph{Testing datasets}: Two datasets are used for testing: AFLW2000~\cite{zhu2016face} and BIWI~\cite{fanelli2013random} datasets.
The AFLW2000 dataset consists of the first $2000$ images in AFLW~\cite{koestinger2011annotated}, which collects face images from the internet.
It is a test set considering different identities, ambience, and quality, as well as large head poses.
BIWI is a dataset of $24$ video sequences. They record $20$ persons rotating across large pose degrees ($±75º$, $±60º$ and $±50º$ for yaw, pitch and roll respectively) about one metre in front of a Kinect V2 camera. $15000$ frames are sampled from those videos.

\emph{Evaluation protocol}:  We follow the protocol of Hopenet~\cite{ruiz2018fine} which is also a landmark-free head pose estimation method. It is trained on the 300W-LP dataset and tested on the two datasets: the AFLW2000 and the BIWI datasets.
When testing on BIWI, MTCNN is used as face detector.

\vspace{-0.5cm}
\subsection{Benchmark Results}
\label{results}
We compare RankPose with the following state-of-the-art methods:
Dlib~\cite{kazemi2014one}, 3DDFA~\cite{zhu2016face}, FAN~\cite{bulat2018super}, KEPLER~\cite{kumar2017kepler},
Two-Stage~\cite{lv2017deep}, HopeNet~\cite{ruiz2018fine}, FSA-Net~\cite{yang2019fsa}, and QuatNet~\cite{hsu2018quatnet}.
Brief description of these methods can be found in the related work section~\ref{sec:related}.

\begin{table}
\caption{MAE Comparison on AFLW2000.}
\label{AFLW2000}
\footnotesize
\begin{tabular}{ p{2cm}||p{2cm} p{2cm} p{2cm}||p{2cm}  }
 \hline
 Method  & Yaw  & Pitch & Roll &Avg. MAE   \\
 \hline
 Dlib~\cite{kazemi2014one}    & 23.1 & 13.6  & 10.5 &15.8   \\
 3DDFA~\cite{zhu2016face}   & 5.40 & 8.53  & 8.25 &7.39   \\
 FAN~\cite{bulat2018super}     & 6.36 & 12.3  & 8.71 &9.12   \\
 HopeNet~\cite{ruiz2018fine} & 6.47 & 6.56  & 5.44 &6.16   \\
 FSA-Net~\cite{yang2019fsa} & 4.50 & 6.08  & 4.64 &5.07   \\
 QuatNet~\cite{hsu2018quatnet} & 3.97 & 5.62  & 3.92 &4.50   \\
 RankPose (ours)& \textbf{2.99} & \textbf{4.75}  & \textbf{3.25} &\textbf{3.66}   \\
 \hline
\end{tabular}
\end{table}

Table~\ref{AFLW2000} compares RankPose with other state-of-the-art methods on the AFLW2000 dataset.
Following previous works, the mean absolute error (MAE) is computed as evaluation metric.
We report the predicted MAE of yaw, pitch, and roll angles of all methods in the second, third, and forth columns respectively.
The average value of MAE of $3$ angles is reported in the last column.
As expected, deep learning-based approaches (3DDFA, FAN, HopeNet, FSA-Net, QuatNet and RankPose) perform better than the method based on
the regression tree (Dlib).
We can observe that landmark-free methods (HopeNet, FSA-Net, QuatNet, and RankPose) outperform landmark-based methods (Dlib, 3DDFA, and FAN).
Our proposed RankPose outperforms all the comparison methods in all dimensions significantly.
Remarkably, RankPose improves the current SOTA result (achieved by QuatNet) from 4.50 to 3.66, which is improvement of $18.66\%$.

\begin{figure}
\centering
\subfigure{\includegraphics[width=0.43 \textwidth]{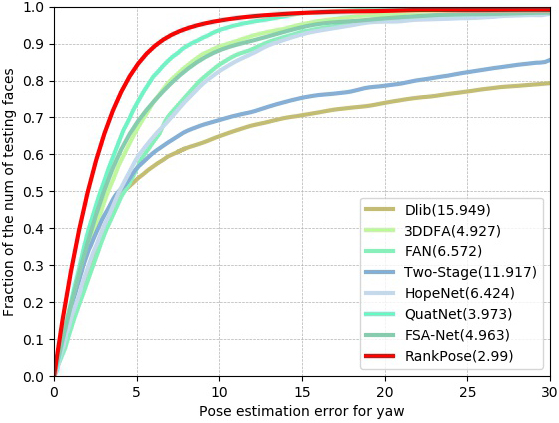}}
 ~~
\subfigure{\includegraphics[width=0.43 \textwidth]{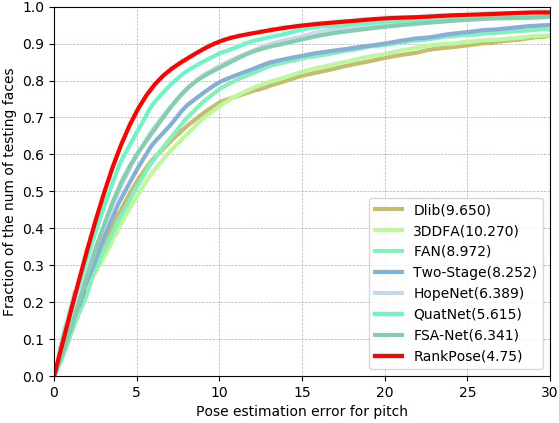}}

\subfigure{\includegraphics[width=0.43 \textwidth]{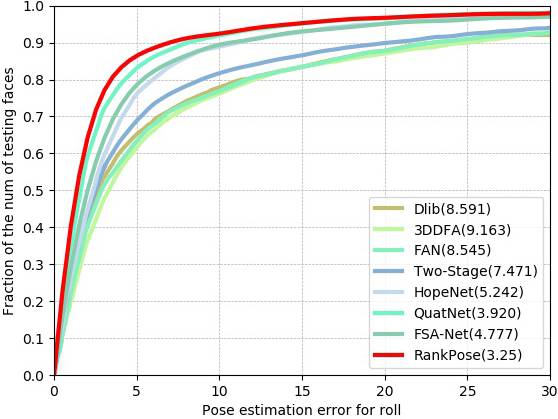}}
\setlength{\belowcaptionskip}{-8pt}
\caption{Error distribution in terms of yaw, pitch and roll.}
\label{fig:error_dist}
\end{figure}

Error distribution in terms of yaw, pitch and roll is plotted in figure~\ref{fig:error_dist}.
The cumulative error curves of RankPose are much skewer than other comparison methods.
There are significant margins between the curves of RankPose and that of other schemes.
Using our method, most test samples ($98.7\%$) are with predicted yaw error less than $15\degree$;
$97.25\%$ are with pitch error less than $20\degree$;
$97.15\%$ are with roll error less than $20\degree$.

\begin{table}
\caption{MAE Comparison on BIWI.}
\label{BIWI}
\footnotesize
\begin{tabular}{ p{2cm}||p{2cm} p{2cm} p{2cm}||p{2cm}  }
 \hline
 Method  & Yaw  & Pitch & Roll &Avg. MAE    \\
 \hline
 Dlib~\cite{kazemi2014one}    & 16.8 & 13.8  & 6.19 &12.2   \\
 3DDFA~\cite{zhu2016face}   & 36.2 & 12.3  & 8.78 &19.1   \\
 FAN~\cite{bulat2018super}     & 8.53 & 7.48  & 7.63 &7.89   \\
 KEPLER~\cite{kumar2017kepler} & 8.80 & 17.3  & 16.2 &13.9   \\
 Two-Stage~\cite{lv2017deep} & 9.49 & 11.34 & 6.00 &8.94   \\
 HopeNet~\cite{ruiz2018fine} & 5.17 & 6.98  & 3.39 &5.18  \\
 FSA-Net~\cite{yang2019fsa} & 4.27& 4.96  & 2.76 &4.00  \\
 QuatNet~\cite{hsu2018quatnet} & 4.01 & 5.49  &2.94 &4.15  \\
 RankPose (ours)& \textbf{3.59} & \textbf{4.77}  & \textbf{2.76} &\textbf{3.71}  \\
 \hline
\end{tabular}
\end{table}

Table~\ref{BIWI} compares RankPose with other state-of-the-art methods on the BIWI dataset.
The characteristics of the training set and BIWI are quite different.
Similar to the result on the AFLW2000 dataset, landmark-free methods perform better than landmark-based methods.
Our RankPose outperforms other comparison schemes significantly showing that it generalises better across deferent domains.
Specifically,  RankPose improves the current SOTA result (achieved by FSA-Net) from 4.0 to 3.71, which is $7.25\%$ improvement.

\subsection{Ablation studies}

\begin{table}
\caption{Ablation studies of different loss functions $\&$ output transformations.}
\label{ablation}
\footnotesize
\begin{tabular}{ p{3.5cm}|p{0.6cm} p{0.6cm} p{0.6cm} p{0.6cm}|p{0.6cm} p{0.6cm} p{0.6cm} p{0.6cm} }
 \hline
 Dataset  & \multicolumn{4}{|c|}{AFLW2000} & \multicolumn{4}{|c}{BIWI}  \\
 \hline
 Method    & Yaw  & Pitch &Roll &Avg. MAE &Yaw &Pitch &Roll &Avg. MAE  \\
 MSE       &3.52  &6.56	  &4.49	&4.85 &4.70	&10.18	&3.36	&6.08\\
 MSE+Cosine  &3.5	  &5.58	  &4.24	&4.44 &4.42	&6.75	&3.38	&4.85\\
 MSE+Arccos &3.51  &5.52	  &4.18	&4.40 &4.77	&5.60	&3.08	&4.48\\
 Ranking loss+MSE &2.90	&5.13 &3.68	&3.91 &3.73	&5.69	&2.79	&4.07\\
 Ranking loss+MSE + Cosine &2.96	&4.70	&3.37	&3.68 &3.64	&4.8 &2.73 &3.72 \\
 Ranking loss+MSE + Arccos  &2.99 &4.75	&3.25	&3.66	&3.59	&4.77	&2.76	&3.71\\
\hline
\end{tabular}
\end{table}

We conduct ablation studies to understand the influence of individual components.
Results are reported in Table~\ref{ablation}.
Two different loss functions: 'MSE' (Equ.~\ref{mse_loss}) and 'ranking loss+MSE' (Equ.~\ref{loss}) are studied;
two different transformation schemes: Cosine transformation (Equ.~\ref{cosine}) and Arccos transformation (Equ.~\ref{arccos}) are compared.
We first compare a baseline model ('MSE') without any transformation with a model using Cosine transformation ('MSE+Cosine') and a model using Arccos transformation ('MSE+Arccos').
Models with transformations show consistently higher performance.
Particularly, the model with Arccos transformation performs the best.
We see the transitive property also applies to the models with ranking loss.
We next compare models ('MSE'/'MSE+Cosine'/'MSE+Arccos') without ranking loss and models ('Ranking loss+MSE'/'Ranking loss+MSE + Cosine'/'Ranking loss+MSE + Arccos') with ranking loss.
Nevertheless, it is most indubitable here that ranking loss+MSE always outperforms MSE alone by a large margin.
The absence of Cosine/Arccos transformation does not affect it being state-of-the-art.

\section{Conclusion}
\label{sec:conclusion}

In this paper, we have discussed the challenges encountered by current head pose estimation methods.
We propose to transform head pose representation learning to a bounded angle space to boost prediction accuracy.
More importantly, we propose a model named RankPose to learn generalised abstract landmarks with ranking loss. With these improvements, we train a model which achieves state-of-the-art results on public benchmarks. Hereby, we deliver our work in a hope that deeper understanding of head pose estimation will arise in future discussions.

\bibliography{egbib}
\end{document}